\documentclass[11pt,a4paper]{article}
\usepackage[hyperref]{emnlp2018}
\usepackage[british]{babel} 
\usepackage[utf8]{inputenc}
\usepackage{times}
\usepackage{url}
\usepackage{amsmath, amsfonts, amssymb, latexsym, textcomp}
\usepackage{graphicx}
\usepackage{booktabs}
\usepackage{physics}
\usepackage{color}
\usepackage{xspace}
\usepackage{listings, algorithm}
\usepackage[noend]{algpseudocode}
\usepackage{CJKutf8} 
\usepackage{multirow, tabularx, makecell}
\usepackage{verbatim}
\usepackage{subfig, caption, float}
\usepackage{flushend,placeins} 
\usepackage[shortcuts]{extdash}

\aclfinalcopy 

\newcommand\ACTIS{ACT\=/ENC\xspace}
\newcommand\ACTOS{ACT\=/DEC\xspace}

\title{Learning to Segment Inputs for NMT\\ Favors Character-Level Processing}

\author{Julia Kreutzer$^{1}$\Thanks{ Work was done while interning at Amazon, Berlin.}  \and Artem Sokolov$^{1,2}$\\
$^{1}$Computational Linguistics, Heidelberg University, Germany \\
$^{2}$Amazon Research, Germany\\
{\tt \small kreutzer@cl.uni-heidelberg, artemsok@amazon.com}
}

\newcommand{\x}{\mathbf{x}}
\newcommand{\y}{\mathbf{y}}
\newcommand{\oo}{\mathbf{o}}
\newcommand{\s}{\mathbf{s}}

\date{}

\begin{document}
\maketitle
\begin{abstract}
    Most modern neural machine translation (NMT) systems rely on presegmented
    inputs. Segmentation granularity importantly determines the input and
    output sequence lengths, hence the modeling depth, and source and target
    vocabularies, which in turn determine model size, computational
    costs of softmax normalization, and handling of out-of-vocabulary words. However,
    the current practice is to use static, heuristic-based segmentations that are fixed before
    NMT training. 
    This begs the question whether the chosen segmentation is optimal for the translation task.
    To overcome suboptimal segmentation choices, we present an algorithm for dynamic segmentation
    based on \citep{Graves:16}, that is trainable end-to-end and driven by
    the NMT objective.  In an evaluation on four translation tasks
    we found that, given the freedom to navigate between different segmentation
    levels, the model prefers to operate on (almost) character level, providing
    support for purely character-level NMT models from a novel angle.
\end{abstract}

\section{Introduction}

Segmentation of input sequences is an essential preprocessing step for neural
machine translation (NMT) and has been found to have a high positive impact on
translation quality in recent WMT shared task evaluations \citep{WMT:16,
WMT:17}. This success can be explained statistically, since shorter segments
are beneficial for reducing sparsity: They lower the type-to-token ratio, decrease the
number of out-of-vocabulary (OOV) tokens and singletons, which in turn improves
the coverage of unseen inputs. 
Two subword segmentation methods are presently the state-of-the-art in NMT: the
\emph{byte-pair encoding} (BPE), that starts with a dictionary of single
characters and iteratively creates a new entry from the two currently most
frequent entries~\citep{Gage:94, SennrichETAL:16}, and a similar,
likelihood-based, \emph{wordpiece} (WP) model by \citet{schuster12}.

While being empirically more successful than word-based NMT, both BPE and WP
are preprocessing heuristics, they do not account for the translation task or
the language pairs at hand (unless applied to both sides jointly), and
require additional preprocessing for languages that lack explicit word
separation in writing.  Being used in a pipeline fashion, they make it
impossible for an NMT system to resegment an unfavorably presplit input
and require consistent application of the same segmentation model during
testing, which adds an integration overhead and contributes to the `pipeline jungles'
in production environments~\citep{SculleyETAL:15}.

On the other extreme from word-based NMT models lie purely character models.
Their advantages are smaller vocabularies, thus smaller embedding and output
layers, allowing for more learning iterations within a training time budget to
improve generalization~\citep{HofferETAL:17}, and no preprocessing requirements.
At the same time, longer input sequences aggravate known optimization problems
with very large depths of time-unrolled RNNs~\citep{HochreiterETAL:01} and may
require additional memory for tracking gradients along the unrolling steps. 

In this work, we pose the following question: \textbf {what would the 
input segmentations look like if the NMT model could decide on them
dynamically?} Instead of heuristically committing to a fixed (sub)word- or
character-segmentation level prior to NMT training, this
would allow segmentation for each input to be driven by the training objective
and avoid solving the trade-offs of different levels by trial and error.  
To answer this question, we endow an NMT model with the capacity of adaptive
segmentation by replacing the conventional lookup embedding layer with a `smart
embedding' layer that sequentially reads input characters and dynamically
decides to group a block of them into an output embedding vector, feeding it to
the upstream NMT encoder before continuing with the next block (with an
optional reverse process on the target side). To signal that a block of
characters, encoded as an embedding vector, is ready to be fed upstream, we use
accumulated values of a scalar halting unit~\citep{Graves:16}, which learns when 
to output this block's embedding. It simultaneously affects
weighting probabilities of intermediate output vectors that compose the
output embedding. Thanks to this weighting, our model is fully
differentiable and can be trained end-to-end. Similarly to BPE, it has a
hyper-parameter that influences segmentation granularity, but in contrast to
BPE this hyper-parameter does not affect the model size.  While we evaluate our on-the-fly
segmentation algorithm on RNN-based NMT systems, it is transferable to
other NMT architectures like CNN-based~\citep{GehringETAL:17} or Transformer
models~\citep{VaswaniETAL:17}, since it only replaces the input embedding layer.

Empirically, we find a strong preference of such NMT models to operate on segments that are
only one to a few characters long. This turns out to be a reasonable choice, as in our
experiments character-level NMT systems of smaller or comparable size were 
able to outperform word- and subword-based systems, which corroborates
results of \citet{ChungETAL:16, ChungETAL:17}. 
Given this finding and the unique advantages of
character-level processing (no pipelining, no tokenization, no additional
hyperparameters, tiny vocabulary and memory, and robustness to spelling
errors~\citep{LeeETAL:17}), 
we hope that character-level NMT, and in general character-level sequence-to-sequence learning, will receive more attention from researchers.

Note that, although our character-based models outperform (sub)word-based ones
with similar architectures on some datasets, we are not seeking to establish a new state-of-the-art
in NMT with our model. Our goal is to isolate the effects of segmentation on quality by introducing a flexibility-enhancing research tool.
Therefore, in the comparisons between (sub)word- and
character-based models we purposely avoided introducing changes to our baseline RNN NMT architecture beyond upgrading the embedding layer. 

\section{Related Work}\label{sec:related}
To tackle the OOV problem in word-level models, \citet{LuongManning:16}
proposed a hybrid model that composes unknown words from characters both on encoder and decoder side. 
While their approach relies on given word boundaries, 
they report a purely character-based baseline performing as well as a
word-based model with unknown word replacement, but taking three months to train,
which seems to have cooled off the NMT community in investigating fully
character-based models as an alternative to (sub)word-based ones. Unlike \citep{LuongManning:16}, we found that despite the
training speed being slower than for (sub)word vocabularies, it is possible to
train reasonable character-level models within a few weeks.

To combine the best of both worlds, \citet{ZhaoZhang:16} proposed hierarchical en-/decoders that receive
inputs on both word- and character-level.
The encoder learns a weighted recurrent representation of each word's characters and the
decoder receives the previous target word and predicts characters until a
delimiter is produced.  
Similar to our work, they find improvements over BPE models. 
The idea to learn composite representations of blocks of characters is
similar to ours, but their approach requires given word boundaries,
which our model learns on-the-fly.

\citet{ChungETAL:16} combined a standard subword-level encoder with a
two-layer, hierarchical character-level decoder. The decoder has gating units
that regulate the influence of the lower-level layer to the higher-level one,
hence fulfilling a similar purpose as our halting unit. This model outperforms a
subword-level NMT system, and achieves state-of-the-art on a subset of
WMT evaluation tasks. 
While not requiring explicit segmentation on the target side, the model still relies on given source segmentations.

Finally, \citet{LeeETAL:17} proposed a 
fully character-level NMT model. They mainly address 
training speed, which \citet{LuongManning:16} identified as a
problem, and introduce a low-level convolutional layer over character
embeddings to extract information from variable-length character n-grams for 
higher-level processing with standard RNN layers. In this way, 
overlapping segments are modelled with a length depending on the convolutional filters. 

Perhaps closest to our work is~\citep{ChungETAL:17}, 
where each layer of a hierarchical RNN encoder is updated at different rates, with the
first layer modelling character-level structures, the following modelling
sub(word)-level structures. They introduce a binary 
boundary detector, similar to our halting unit, that triggers feeding of a representation to the next
level, so that latent hierarchical structures without explicit boundary
information are learnt. Unlike our fully-differentiable model, such discrete decisions of the boundary detector 
prohibit end-to-end differentiability, forcing a recourse to the biased straight-through estimator~\citep{BengioETAL:13}.
On the other hand, while our model relies on a to-be-tuned
computation time penalty, 
\citet{ChungETAL:17} do not impose constraints on the number of boundaries. 

Concurrently to our work, \citet{CherryETAL:18} adapt hierarchical multi-scale RNNs \citep{ChungETAL:17} to NMT and compare them to several compression algorithms for character-based NMT. Similar to our work, they focus on the encoder and come to the same conclusion: deep recurrent models at character level work surprisingly well.

\section{Jointly Learning to Segment and Translate} 

Instead of committing to a single segmentation before NMT model training, we
propose to learn the segmentation-governing parameters along with the usual network parameters in a end-to-end differentiable manner. 
With this approach, we get rid of pipelining and pre-/postprocessing, and can adaptively segment arbitrary inputs we encounter during training or testing. 
Our segmentations are context-dependent, i.e.\ the same substring can be
segmented into different parts in different contexts. Being able to
smoothly interpolate between word-based and character-based models we allow the
model to find a sweet spot in between.

We extend the \emph{Adaptive Computation Time} (ACT) paradigm introduced
by \citet{Graves:16}, where a general RNN model is augmented with a scalar
halting unit that decides how many recurrent computations are spent on each
input. For segmentation, we use the halting unit to decide how many inputs
(characters) a segment consists of. The output of the ACT module can thus be
thought of as an `embedding' vector for a segment that replaces the classic
lookup embedding for (sub)words in standard NMT models.  While our model can in principle use
larger units as elementary inputs,
we will focus on character inputs to be able to model the composition of arbitrary segments. That means that we only add a small amount of parameters to a basic character-based
model, but explicitly model higher-level merges of characters into subwords. 

\subsection{ACT for Dynamic Depth}\label{sec:act}
\begin{figure}[t]
\centering
\includegraphics[width=0.4\textwidth]{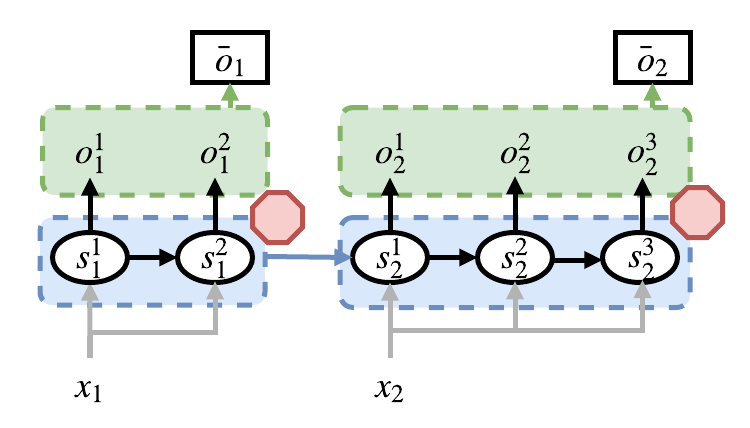}
\caption{\citet{Graves:16}'s ACT. Each input is repeatedly fed (gray
    arrows) into the recurrent functions (ellipses) that produce intermediate
    states and outputs for each input step. When the halting unit halts
    (red stop sign), intermediate states are summarized (weighted mean,
    blue), as well as outputs (green). These summaries, respectively, form the input to the
    following cell's state or are output from the network.}
\label{fig:act-sketch}
\end{figure}

Here we summarize the ACT model \citep{Graves:16}, depicted in
Figure~\ref{fig:act-sketch}. It is applicable to any recurrent architecture
that transforms an input sequence $\x=(x_1,\dots,x_{T})$ into outputs $\oo=(\bar
o_1,\dots,\bar o_{T})$ via computing a sequence of states
$\s=(s_1,\dots,s_T)$ through a state transition function~$\mathcal{S}$ on an
embedded input $E x_t$ and a linear output projection defined by
matrix $W_o$ and bias~$b_o$:
\begin{align}
	s_t=\mathcal{S}(s_{t-1}, E x_t), \qquad 
    o_t=W_o s_t+b_o \label{eq:act-output}
\end{align}

Instead of stacking multiple RNN layers in $\mathcal{S}$ to achieve increased
complexity of an RNN network, the ACT model dynamically decides on the number
of necessary recurrent steps (layers) for every input $x_t$. This saves
computation on easy inputs, while still being able to use all of the processing power 
on hard inputs before emitting outputs. Concretely, an ACT cell performs
an arbitrary number of internal recurrent applications of $\mathcal{S}$ for each input $x_t$:\footnote{In \cite{Graves:16} repeated inputs are augmented with a binary flag, which we ignore here for the sake of simplicity.}
\begin{align}
	s^n_t&=\begin{cases}\mathcal{S}(\bar{s}_{t-1}, E x_t), \text { if } n=1\\ \mathcal{S}(s^{n-1}_t, E x_t), \text{ otherwise}\end{cases}\label{eq:act}
\end{align}
The total number of internal steps is $N(t)=\min\{n':\sum^{n'}_{n=1}h^n_t\geq 1-\epsilon\}$, where $\epsilon \ll 1$ and $h^n_t$ is the scalar output of sigmoid halting unit, 
\begin{equation}
    h^n_t=\sigma(W_h s^n_t+b_h). \label{eq:halting_unit}
\end{equation}
Once halted, the final output $\bar o_t$ and state $\bar s_t$ (which is fed to
the next ACT step in~\eqref{eq:act}) are computed as weighted means of
intermediate outputs and states:
\begin{align}
    \bar{s}_t=\sum_{n=1}^{N(t)} p_t^n s_t^n, \qquad
    \bar{o}_t=\sum_{n=1}^{N(t)} p_t^n o_t^n \label{eq:mean-output}
\end{align}
 where probabilities $p^n_t$ are defined as
\begin{align}
 p^n_t =\begin{cases} 
            R(t), \text { if }n=N(t)\\
            h^n_t, \text{ otherwise}
        \end{cases}\label{eq:halting_prob}
\end{align}
and remainders
$R(t) = 1 - \sum_{n'=1}^{N(t)-1}h^{n'}_t.$
Finally, to prevent the network from pondering on an input for too long, the
remainder $R(t)$ is added as a penalty to the RNN training loss (usually
cross-entropy (XENT)) with a weight~$\tau$:
\begin{align}
    L_{\text{ACT}}=L_{\text{XENT}}+\tau R(t).\label{eq:act-loss}
\end{align}
Thanks to \eqref{eq:mean-output}, the model is deterministic and differentiable. 

\subsection{ACT for Dynamic Segmentation}\label{sec:actsegment}
Dynamic segmentation can either be applied on the source side or the target
side or on both. We focus on an ACT-encoder (\ACTIS) with dynamic segmentation
for the source side, and describe an ACT-decoder model for the target side (\ACTOS), which dynamically segments outputs by compounding output characters, in Appendix~A.  

\begin{figure}[t]
\centering
\includegraphics[width=0.4\textwidth]{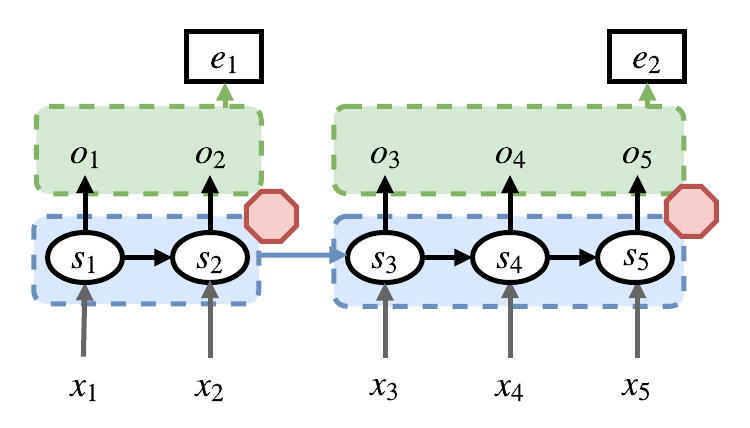}
    \caption{Diagram of the \ACTIS encoder. Note the differences to the original ACT model: An input is here read on \emph{every} internal recurrent iteration (gray arrows) and the halting unit (red stop sign) is repurposed to trigger feeding of an encoded embedding vector of a block of characters to the upstream NMT layers.}
\label{fig:act-enc-sketch}
\end{figure}

\paragraph{Segmenting Encoder.}\label{sec:encoder}
We now describe how to use the ACT paradigm to enhance an encoder for dynamic segmentation on the source side (\ACTIS).
We reuse the idea of halting units, mean field updates and $\tau$-penalized training objective, but 
instead of learning how much computation is needed for each
atomic input, we learn how much computation to allow for an aggregation of
atomic inputs, i.e. one segment.

The input to an \ACTIS cell is a sequence of one-hot-encoded
characters
$\x = (x_1, \dots,
x_{T_\x})$. The \ACTIS, depicted in Figure~\ref{fig:act-enc-sketch}, receives one
input $x_t$ at a time and decides whether to halt or not. In the case of no
halting, the cell proceeds reading more inputs; if it halts, it produces
an output `embedding' $\bar o$ of a block of characters read so far, and the
cell resets for reading the next block.  The sequence of the output
embeddings $\oo = (\bar o_1, \dots, \bar o_{T_\oo})$ is then fed to upstream standard (possibly
bidirectional) NMT encoder layers, replacing the usual, one-hot encoded, (sub)word lookup embeddings.
The length of $\oo$ is variable: The more frequently \ACTIS halts,
the more embeddings are generated. In extreme cases, it can generate one
embedding per input ($T_\oo = T_\x$) or just one embedding for the full sequence of
inputs ($T_\oo=1$). 

\begin{algorithm}[t]
\begin{algorithmic}[1]
    \Require{Weights $W_{o},b_{o},W_{h},b_{h}$, transition function $\mathcal{S}$, embeddings $E_{src}$, inputs $\x = (x_1, \dots, x_{T_\x})$}
    \Ensure{Outputs $\oo=(\bar{o}_1,\dots, \bar{o}_{T_\oo})$, remainder $R$}
\State{$\oo = [\;]$} \Comment{empty sequence} 
\State{$R = 0$}  \Comment{init remainder} 
    \State{$\bar{s} = \mathbf{0}$, $\bar{o} = \mathbf{0}$} \Comment{init mean state and output}
\State{$H = 0$} \Comment{init halting sum} 
\State{$s_0 = \mathbf{0}$} \Comment{init state}
\For{$t = 1 \dots T_\x$} \Comment{loop over inputs}
	\State{$s_t = \mathcal{S}(s_{t-1}, E_{src}\,x_t)$} \label{actis:state} \Comment{new state}
	\State{$o_t = W_{o}s_t + b_{o}$} \label{actis:out} \Comment{new output}
	\State{$h_t = \sigma(W_{h} s_t + b_{h})$} \label{actis:score} \Comment{halting score}
	
    \State{$f = [\![H+h_t \geq 1-\epsilon ]\!]$} \Comment{halting flag}
    \State{$p_t = (1-f) \, h_t + f \, (1-H)$} \label{actis:prob} \Comment{halting probability}
    \State{$H = H + h_t$} \Comment{update halting sum}
    \State{$\bar{s} =\bar{s} + p_t s_t$} \Comment{mean state}
    \State{$\bar{o} = \bar{o}+ p_t o_t$} \Comment{mean output}
	\State{$R = R+(1-f)\,h_t$} \label{actis:penalty} \Comment{increment remainder}
	\If{$f$} \label{actis:halt}
	\State{$\oo = \oo^\frown [\bar{o}]$} \Comment{append output} \label{actis:expand}
	\State{$s_t = \bar{s}$} \label{actis:overwrite}\Comment{overwrite for next step}
	\State{$\bar{s} = \mathbf{0},\, \bar{o}= \mathbf{0},\, H = 0$}
	\EndIf
\EndFor
\State{$R = \frac{1-R}{t}$}\label{actis:normalize} \Comment{normalize remainder}
\end{algorithmic}
\caption{\ACTIS}
\label{alg:actis}
\end{algorithm}

In more detail, \ACTIS implements
the pseudocode given in Algorithm \ref{alg:actis}.  Let
$\mathcal{S}(s_{t-1}, i_{t})$ be any recursive computation function (in this work we use GRUs) 
of an RNN that receives a hidden state $s_{t-1}$ and an input vector
$i_{t}$ at time step $t$ and computes the new hidden state $s_t$. In line
\ref{actis:state} this function is computed on the regular previous state or,
if there was a halt in the previous step (line~\ref{actis:halt}), on the mean
state vector~$\bar{s}$ that summarizes the states of the previous segment
(line~\ref{actis:overwrite}, cf.~\eqref{eq:mean-output}, 1st eq.).  Per-step outputs $o_t$ are computed from the
hidden states $s_t$ with a feed-forward layer (line~\ref{actis:out},
cf.~\eqref{eq:act-output}, 2nd eq.).  A sigmoid halting unit computes a halting score
in each step (line~\ref{actis:score}, cf.~\eqref{eq:halting_unit}).  The
halting probability for step $t$ is either the halting score $h_t$ or the
current value of remainder $1-H$ to ensure that all halting probabilities
within one segment form a distribution (line~\ref{actis:prob},
cf.~\eqref{eq:halting_prob}). $\epsilon$ is set to a small number to allow
halting after a single step. 
Whenever the
model decides to halt, an output embedding~$\bar o$ is computed as a weighted mean of the
intermediate outputs of the current segment (line~\ref{actis:expand},
cf.~\eqref{eq:mean-output}, 2nd eq.).  The weighted mean on the one hand serves the
purpose of circumventing stochastic sampling, on the other hand can be
interpreted as a type of intra-attention summarizing the intermediate states
and outputs of the segment.
The halting scores from each step are accumulated (line~\ref{actis:penalty}) to
penalize computation time as in~\eqref{eq:act-loss}. The hyperparameter $\tau$ here
controls the segment length: The higher its value, the more preference will
be given to smaller remainders, i.e.\ shorter segments.
We introduce an additional normalization by input length (line~\ref{actis:normalize}), such that longer sequences will be allowed
more segments than shorter sequences. 
This implementation exploits
the fact that \ACTIS outputs are weighted means over time steps and
updates them incrementally.  The algorithm allows efficient minibatch
processing by maintaining a halting counter that indicates which embedding each current
intermediate output in the batch contributes to. Incremental updates of
embeddings and states are achieved with masks depending on the halting
position. 

\paragraph{Segmenting Decoder.} 
We also implemented a similar segmenting decoder, \ACTOS (see Appendix~A), that
`transcribes' vectors emitted by an NMT decoder into a variable number of
characters. Our preliminary experiments with both adaptive input \emph{and}
output segmentation capabilities did not improve over using only \ACTIS
with a standard character-level NMT decoder, so in this paper we report only
results of the latter configuration. 

\paragraph{Comparison to the Original ACT.} 
While our \ACTIS reuses the ideas
of halting units, mean field updates and $\tau$-penalized training objective,
it has the following differences to the original ACT: First of all, our model has a different
purpose and addresses segmentation, not the alignment of pondering time to
input complexity.  Instead of learning how much computation is needed for each
atomic input, we learn how much computation to allow for an aggregation of
atomic inputs.  Second, it has a different halting behavior: \ACTIS allows
multiple halts per sequence, not only one per character ($N(t)$ is no longer a
function of~$t$).  More similar to ACT, our segmenting decoder \ACTOS
(Appendix~A) has one halt per input element, but can generate
arbitrarily many output characters per input.

\section{Experiments}\label{sec:exp}
We reimplemented the Groundhog RNN encoder-decoder model with attention
by~\citet{BahdanauETAL:15} in \texttt{MxNet}
\texttt{Gluon} 
to allow for dynamic computation graphs. To cover a
wide range of linguistic diversity, we report results on four language
directions and domains, for \mbox{word-,} subword-, character-level and
\ACTIS segmentation: German-to-English TED talks, Chinese-to-English web pages, Japanese-to-English scientific abstracts and French-to-English news.
Table~\ref{tab:data} gives an overview of the datasets.  

\begin{table}[h]
\centering
\resizebox{\columnwidth}{!}{
\begin{tabular}{lllrrr}
    \toprule
    \textbf{Data} & \textbf{Domain} & \textbf{Lang} & \textbf{Train} & \textbf{Dev} & \textbf{Test} \\
    \midrule
    IWSLT & TED talks      & de-en & 153,352    & 6,970 & 6,750\\
    CASIA & web            & zh-en & 1,045,000  & 2,500 & 2,500\\
    ASPEC & sci. abstracts & ja-en & 2,000,000  & 1,790 & 1,812\\
    WMT   & news           & fr-en & 12,075,604 & 6,003 & 3,003\\
    \bottomrule
\end{tabular}
}
\caption{Domain, language pairs and number of parallel sentences per split for the used datasets.}
\label{tab:data}
\end{table}

\paragraph{Preprocessing and Evaluation.}
The IWSLT data is split
and processed as in~\citet{BahdanauETAL:17}; 
since it comes pretokenized and lowercased\footnote{\url{https://github.com/rizar/actor-critic-public/tree/master/exp/ted}}, models are evaluated with
tokenized, lowercased BLEU (using \texttt{sacrebleu}\footnote{\url{https://pypi.org/project/sacrebleu}}) and chrF-score on character
bigrams~\citep{Popovic15}. 
For WMT, we used the 2014 dataset prepared by~\citet{BahdanauETAL:15}\footnote{
\url{http://www-lium.univ-lemans.fr/~schwenk/cslm_joint_paper/}}, additionally
filtering the training data to include only sequences of a lengths 1 to 60,
and models are evaluated with cased BLEU and chrF (\texttt{sacrebleu}, with the ``13a'' tokenizer).  
The CASIA and ASPEC data are, respectively, from the 2015 China Workshop on MT (CWMT), used without additional pre-processing, and 
from the WAT~2017 SmallNMT shared task, 
pretokenized with WP. Both datasets have BPE and WP vocabularies of around 16k for each side, and we report cased BLEU and chrF on them.

\paragraph{Hyperparameters.} 
All models are trained with Adam~\citep{KingmaBa:14} and a learning rate of 0.0003, halved whenever the validation score (tokenized BLEU) has not increased for 3 validations. Training stopped when the learning rate has been decreased 10 times in a row. Models are validated every 8000 training instances.
All models use recurrent cells of size 1000 for the decoder, with a bidirectional encoder of size 500 for each direction, input and output embedding of size 620, and the attention MLP of size 1000, all following~\citep{BahdanauETAL:15}. 
When multiple encoders layers are used for character-based models, they are all bidirectional~\citep{ChenETAL:2018} with attention on the uppermost layer. 
The ACT layer for \ACTIS models has size 50 for IWSLT, CASIA and ASPEC, and 25 for WMT (we picked the ACT size over $\{25, 50, 75, 100, 150\}$). 
The word-based models on IWSLT and WMT have a vocabulary of 30k for each side. BPE models have separate 15k vocabularies for IWSLT, and a joint 32k vocabulary for WMT.
For IWSLT, CASIA and ASPEC all characters from the training data were included in the vocabularies, 117 (de) and 97 (en), 7,284 (zh) and 166 (en), and 3,212 (ja) and 233 (en), respectively. 
For WMT the vocabularies included the 400 most frequent characters on each side.
\mbox{Word-} and BPE-based models are trained with minibatches of size 80,
character-based models with~40. The maximum sequence length during
training is 60 for \mbox{word-} and BPE-based models, 200 for
character-based models and 150 for \ACTIS, to fit into available memory.
\citet{Graves:16} observed that tuning $\tau$ was crucial for success of
ACT. A suboptimal value of~$\tau$, that in our case influences
possible segment lengths, might make it hard to achieve good performance in
terms of BLEU or chrF. We therefore searched over a range of $\tau$s\footnote{$\{-0.5, 0, 0.001, 0.3, 0.4, 0.45, 0.5, 0.505, 0.55, \\ 0.49, 0.6, 0.7, 0.75, 0.8, 0.85, 0.9, 1.0, 1.5\}$} on the dev sets, keeping
other hyperparameters fixed:
$\tau=1.0$ delivered the highest BLEU score for IWSLT and CASIA, $\tau=0.8$ for WMT and $\tau=0.7$ for \mbox{ASPEC}.
Following~\citet{Graves:16}, we fixed $\epsilon=0.01$ in all the experiments. 
During inference, we
use beam search with a beam size of 5 and length-normalization parameters
$\beta=0.0, \alpha=1.0$~\citep{WuETAL:16}.

\begin{table}
\centering
\resizebox{\columnwidth}{!}{
\begin{tabular}{ll|cc|rll}
\toprule
    \textbf{Data} & \textbf{Model} &  \textbf{BLEU} & \textbf{chrF}  & \textbf{Param} & \textbf{SegLen} & \textbf{TrainTime}\\
\midrule
    \multirow{4}{*}{\makecell{IWSLT\\de-en}} 
    & Word & 22.11 & 0.44 & 80.5M & 4.66 & 23h \\ 
    & BPE  & 25.38 & 0.49 & 46.5M & 4.09 & 20h \\
    & Char & 22.63 & 0.46 & 13.4M & 1.00 & 1d22h \\ 
    & \ACTIS & 22.67 & 0.46 & 13.5M & 1.88 & 9d21h \\ 
    \midrule
\multirow{3}{*}{\makecell{CASIA\\zh-en}} 
    & BPE & 10.59 & 0.37 & 49.9M & 1.72 & 18h \\
    & Char & 12.60 & 0.40 & 21.0M & 1.00 & 10d6h \\
    & \ACTIS & 9.87 & 0.36 & 21.3M & 1.006 & 3d13h \\ 
\midrule
\multirow{3}{*}{\makecell{ASPEC\\ja-en}} 
    & WP & 21.05& 0.53 & 50.0M & 2.07 & 4d4h \\
    & Char & 22.75 & 0.55 & 15.6M & 1.00 & 24d15h \\ 
    & \ACTIS & 15.82 & 0.46 & 15.6M & 1.0007 & 15d4h \\ 
\midrule
\multirow{4}{*}{\makecell{WMT\\fr-en}} 
    & Word & 20.32 & 0.49 & 80.5M & 5.19 & 4d9h \\ 
    & BPE   & 27.02 & 0.55 & 86.0M & 4.05 & 3d23h \\ 
    & Char & 24.25 & 0.53 & 14.1M & 1.00 & 9d \\ 
    & \ACTIS & 13.74 & 0.42 & 14.2M & 1.82 & 13d8h \\
\bottomrule
\end{tabular}
}
    \caption{Evaluation results on respective test sets for 1-layer models, and number of parameters and average source segment lengths on dev sets. Training time to reach stopping criterion is given in (d)ays and (h)ours.}
\label{tab:results}
\end{table}

\paragraph{Results.} 
Table~\ref{tab:results} lists the results for the 
most comparable, 1-layer, configuration.  
BPE models expectedly outperform word-based models, however word-based models are also
outperformed by character-based models.
The picture is similar w.r.t.\ the chrF with even smaller relative differences. 
The \ACTIS model with one unidirectional ACT layer
manages to match the 1-layer bidirectional character-based model on IWSLT. But it does not
reach the results of other models on CASIA and ASPEC, which can be explained
by 
increased complexity of doing simultaneous
segmentation during training on sentences longer than the average sentence length in IWSLT. 
However, the main finding here is that \ACTIS recovers an almost character-level
segmentation (compare the ``SegLen'' column in Table~\ref{tab:results}). On the IWSLT dev set, the average segment length is only 1.88 (with a
maximum of 5 chars per segment).
For \mbox{CASIA} and ASPEC domains and with the larger datasets than IWSLT, the \ACTIS segmentations
becomes more fine-grained: The average segment length is, respectively, just 1.006 and 1.0007 on the dev
set, with a maximum of 2 chars per segment.  Given that
the character model outperforms the model with the BPE/WP segmentation, it is not surprising that \ACTIS converged to the character
segmentation. 

We hypothesize that \ACTIS could not improve over the 1-layer bidirectional character model because of complexity of identifying segments in Chinese and Japanese, unidirectionality of its initial layer, and increased hardness of optimization of character-based models with extra non-linearities~\cite{LingETAL:15}, that causes earlier convergence to poorer minima in many runs.
Similarly for WMT, failing to match the performance of the character model could be caused by harder optimization task on particularly long sentences in the WMT data, and unidirectionality of \ACTIS.
The \ACTIS's segment length on the dev set is 1.82 (max. 6 characters per segment), again close on average to a purely character segmentation.

Inspired by the \ACTIS's recovery of almost character segmentation
and by the competitive performance of pure character-based models, we decided to verify if
the advantage of character-level processing carries over to multiple layers.
Since the character models are much smaller than their word-/BPE-based counterparts, one should allow
multiple layers (consuming the same or less memory) to make up for the difference in number of parameters for fairer
comparison. This also aimed to 
verify whether an 
increased number of non-linearities (one of ACT's benefits \citep{FojoETAL:18}) plays a role.

\begin{table}[t]
\centering
\resizebox{\columnwidth}{!}{
\begin{tabular}{ll|cc|rl}
\toprule
    \textbf{Data} & \textbf{Model} &  \textbf{BLEU} & \textbf{chrF}  & \textbf{Param} & \textbf{TrainTime}\\
\midrule
    \multirow{4}{*}{\makecell{IWSLT\\de-en}} 
    & Word, 4-layer & 24.54 & 0.45 & 97.0M & 1d8h\\
    & BPE, 1-layer  & 25.38 & 0.49 & 46.5M & 20h \\ 
    & Char, 5-layer & \textbf{28.19} & \textbf{0.51} & 26.9M & 3d10h\\ 
    & \ACTIS, 3-layer & 25.10 & 0.49 & 25.6M & 9d7h\\
    \midrule
\multirow{3}{*}{\makecell{CASIA\\zh-en}} 
    & BPE, 3-layer    & 11.01 & 0.38 & 58.9M & 24h\\ 
    & Char, 3-layer   & \textbf{13.43} & \textbf{0.42} & 30.0M & 5d6h\\ 
    & \ACTIS, 2-layer & 10.35 & 0.37 & 21.3M & 10d\\ 
\midrule
\multirow{3}{*}{\makecell{ASPEC\\ja-en}} 
    & WP, 3-layer   & 22.02 & \textbf{0.55} & 61.4M & 4d2h \\ 
    & Char, 1-layer & \textbf{22.75} & \textbf{0.55} & 15.6M & 24d15h\\ 
    & \ACTIS, 1-layer & 15.82 & 0.46 & 15.6M & 15d4h \\
\midrule
\multirow{4}{*}{\makecell{WMT\\fr-en}} 
    & Word, 2-layer  & 21.04 & 0.48 & 94.0M & 4d16h\\ 
    & BPE, 3-layer  & \textbf{27.93} & \textbf{0.56} & 98.0M & 5d3h\\ 
    & Char, 6-layer & 27.23 & 0.55 & 27.6M & 18d13h \\ 
    & \ACTIS, 2-layer & 14.01 & 0.43 & 21.7M & 9d10h \\
\bottomrule
\end{tabular}
}
    \caption{Results on respective test sets after tuning number of encoder layers on the dev set.}
\label{tab:morelayers}
\end{table}

\begin{table*}[t]
\centering
\resizebox{\textwidth}{!}{
\begin{tabular}{ll}
\toprule
\textbf{Ref} & in social groups of animals , the juveniles always look different than the adults .\\
\midrule
\multirow{2}{*}{\textbf{Word}} & in gruppen sozialer tiere sehen die jungtiere immer anders aus als die alttiere . \\
& in groups of social animals , the children are always different from the other than the $<$unk$>$.\\
\midrule
\multirow{2}{*}{\textbf{BPE}} & in gruppen sozialer tiere sehen die jung@@ tiere immer anders aus als die alt@@ tiere .\\
& in groups , in groups , the juveniles are seeing the same animals as well as the animals . \\
\midrule
\multirow{2}{*}{\textbf{\ACTIS}} & in$|$ g$|$ru$|$pp$|$en$|$ s$|$oz$|$ia$|$le$|$r $|$ti$|$er$|$e $|$se$|$he$|$n $|$d$|$ie$|$ j$|$un$|$gt$|$ie$|$re$|$ i$|$m$|$m$|$er$|$ a$|$nd$|$er$|$s $|$au$|$s $|$al$|$s $|$d$|$ie$|$ a$|$lt$|$ti$|$er$|$e $|$.$|$\\ 
& in groups , the juvenile seems to see the different approach than the algaes . \\
\midrule
\multirow{2}{*}{\textbf{Char}} & i$|$n$|$ $|$g$|$r$|$u$|$p$|$p$|$e$|$n$|$ $|$s$|$o$|$z$|$i$|$a$|$l$|$e$|$r$|$ $|$t$|$i$|$e$|$r$|$e$|$ $|$s$|$e$|$h$|$e$|$n$|$ $|$d$|$i$|$e$|$ $|$j$|$u$|$n$|$g$|$t$|$i$|$e$|$r$|$e$|$ $|$i$|$m$|$m$|$e$|$r$|$ $|$a$|$n$|$d$|$e$|$r$|$s$|$ $|$a$|$u$|$s$|$ $|$a$|$l$|$s$|$ $|$d$|$i$|$e$|$ $|$a$|$l$|$t$|$t$|$i$|$e$|$r$|$e$|$ $|$.$|$ \\
& in groups of social animals , the juveniles are still in the alite of the altients .\\
\midrule
\midrule
\textbf{Ref} &  we \&apos;re living in a culture of jet lag , global travel , 24-hour business , shift work . \\
\midrule
\multirow{2}{*}{\textbf{Word}} & wir leben in einer zivilisation mit jet-lag , weltweiten reisen , nonstop-business und schichtarbeit .\\
& we live in a civilization with $<$unk$>$ , global travel , $<$unk$>$ and $<$unk$>$ . \\
\midrule
\multirow{2}{*}{\textbf{BPE}} & wir leben in einer zivilisation mit jet@@ -@@ lag , weltweiten reisen , non@@ sto@@ p-@@ business und sch@@ icht@@ arbeit .\\
& we live in a civilization with a single , a variety of global travel , presidential labor and checking .\\
 \midrule
\multirow{2}{*}{\textbf{\ACTIS}}& w$|$ir$|$ l$|$eb$|$en$|$ i$|$n $|$ei$|$ne$|$r $|$z$|$iv$|$il$|$is$|$at$|$io$|$n $|$m$|$it$|$ j$|$et$|$-la$|$g $|$,$|$ w$|$el$|$tw$|$ei$|$te$|$n $|$re$|$is$|$en$|$ ,$|$ n$|$on$|$st$|$op$|$-bu$|$si$|$ne$|$ss$|$ u$|$nd$|$ s$|$ch$|$ic$|$ht$|$ar$|$be$|$it$|$ .$|$\\
& we live in a civilization with jes lag , worldwide rows , nonstop business and failing . \\ 
\midrule
    \multirow{2}{*}{\textbf{Char}} &  w$|$i$|$r$|$ $|$l$|$e$|$b$|$e$|$n$|$ $|$i$|$n$|$ $|$e$|$i$|$n$|$e$|$r$|$ $|$z$|$i$|$v$|$i$|$l$|$i$|$s$|$a$|$t$|$i$|$o$|$n$|$ $|$m$|$i$|$t$|$ $|$j$|$e$|$t$|$-$|$l$|$a$|$g$|$ $|$,$|$ $|$w$|$e$|$l$|$t$|$w$|$e$|$i$|$t$|$e$|$n$|$ $|$r$|$e$|$i$|$s$|$e$|$n$|$ $|$,$|$ $|$n$|$o$|$n$|$s$|$t$|$o$|$p$|$-$|$b$|$u$|$s$|$i$|$n$|$e$|$s$|$s$|$\\ & $|$u$|$n$|$d$|$ $|$s$|$c$|$h$|$i$|$c$|$h$|$t$|$a$|$r$|$b$|$e$|$i$|$t$|$ $|$.$|$\\ 
& we live in a civilization with jet walk , global journeys , nonstop-business and layering \\
\bottomrule
\end{tabular}
}
\caption{Examples from the IWSLT dev set: segmented sources and greedy translations. Word, BPE and \ACTIS models have 1 encoder layer, and the character model has 5 layers.}
\label{tab:examples-iwslt}
\end{table*}

Table~\ref{tab:morelayers} shows the test results after tuning the number of bidirectional encoder layers, from 1 to 6,
on dev sets. First, we observe the modest parameter number of character models even with multiple layers, that allows them to take advantage of deeper cascades of non-linearities while staying well below the memory budget of (sub)word-based 1-layer models.
Second, we discover that BPE/WP models are outperformed by character-based models with multiple encoder layers on two datasets,
achieving gains of 2.8 BLEU points on IWSLT, 0.7
BLEU on \mbox{ASPEC}, and losing half a point only on WMT (with a minor decrease in chrF), despite having at least
3.5 times fewer parameters.
Such ranking of character- and BPE-based models on WMT might
be explained by much longer sentences in the corpus, compared to IWSLT and
\mbox{ASPEC}, 
since the ability of character and ACT-based models to cover unseen input is
limited by the maximum training sequence length limit (here~200 characters), which on WMT
data crops 30.5\% of sentences.

\paragraph{Translation Analysis.} 
Randomly selected translation examples from the IWSLT dev
set and their segmented sources are given in Table~\ref{tab:examples-iwslt} (more in Table~\ref{tab:examples-wmt} in Appendix~D).
In general, when encountering rare inputs, word-based models fail by producing
the unknown word token ($<$unk$>$), and the BPE-model is able to translate only a more common
part of German compounds (e.g.\ `tiere' $\rightarrow$ `animals'). The character-based models invent
words (`altients', `jes lag') that are similar to strings that they saw during
training and the source. In a few cases they fallback to a language-modeling
regime having attended to the first characters of a corresponding source word:
e.g., instead of translating `reisen' to `journeys', the \ACTIS model
translates it to `rows' (confusing `reisen' to a similarly spelled German
`reihen'), or `layering' instead of `shift work' (confusing `schichten' to the prefix-sharing 
`schichtarbeit'). 
This is confirmed by attention plots in
Figure~\ref{fig:att-char1} in Appendix~C: The model frequently
attends to the correct source word, but mainly to the first characters only;
when adding~4 more layers, the character model develops a behavior
to attend to the first positions of source sentence words, see
Figure~\ref{fig:att-char5}.
Note that \ACTIS segmentations are context-dependent, e.g.\ occurrences of 'tiere' are segmented differently. 

\paragraph{Segmentation Analysis.}
Table~\ref{tab:segs} lists the
most frequent segments produced by 1-layer \ACTIS. For IWSLT, we observe that many segments make sense
statistically (frequent or rare patterns) and linguistically to some extent:
Many of the frequent segments include whitespace (itself a frequent symbol);
2\=/gram segments amongst others include frequent word suffixes (`en',
`in', `er'), but also frequent diphthongs (`ei' and `ie'); 3\=/grams
start with rare characters like `x' and `y' or single dashes; 4\=/grams
combine single characters with whitespaces and double dashes; 5\=/grams
cover numbers, in particular, years.
Importantly, though, since the best test BLEU scores were obtained
by a multi-layer character-based model, the \ACTIS model has done a reasonable job 
in improving over the 
already well-performant
strategy, one character per segment, despite having only a single NMT layer.

For CASIA and ASPEC, \ACTIS converged to a segmentation even closer to pure characters and the longest segments consist of 2 characters. As shown in Table~\ref{tab:segs}, the most frequent 2-grams for CASIA are punctuation marks combined with frequent pronoun \begin{CJK}{UTF8}{min}他\end{CJK} or preposition \begin{CJK}{UTF8}{min}的\end{CJK}, or with the hieroglyph \begin{CJK}{UTF8}{min}明\end{CJK} from a common phrase `[smth.] shows, [that]' (all 4-10k in train), and parts of rare English words.
For ASPEC, it is mostly the Hiragana letter
'\begin{CJK}{UTF8}{min}き\end{CJK}' that starts the segments. While this
letter also occurs as singleton (183 times in the dev set, vs.~52 times as
part of a learned segment), and is rather frequent in the training set (239k), it
is not the most frequent letter. 
See Table~\ref{tab:examples-aspec} in Appendix~D for translation examples for
wordpiece- and character-level models.

\begin{table*}
\centering
\resizebox{0.8\textwidth}{!}{ 
\begin{tabular}{lll}
\toprule
    \textbf{Data} & \textbf{Len} & \textbf{Segments} \\
\midrule
    \multirow{4}{*}{IWSLT} & 2 & en; n\textvisiblespace; er; \textvisiblespace d; ie; e\textvisiblespace; ei; in; \textvisiblespace s; \textvisiblespace w ~~\dots\\
    & 3 & yst; -\textvisiblespace d; xtr; -\textvisiblespace u; 100; xpe; -\textvisiblespace w; xis; -\textvisiblespace e; -ge ~~\dots\\
    & 4 & -{}-\textvisiblespace d; -{}-\textvisiblespace w; -{}-\textvisiblespace s; -{}-\textvisiblespace i; -{}-\textvisiblespace e; -{}-\textvisiblespace u; -{}-\textvisiblespace g; -{}-\textvisiblespace m; -{}-\textvisiblespace a; -{}-\textvisiblespace k ~~\dots\\
    & 5 & 1965\textvisiblespace; 969\textvisiblespace,; 1987\textvisiblespace; 1938\textvisiblespace; 1621\textvisiblespace; 1994\textvisiblespace; 
       1985\textvisiblespace; 1979\textvisiblespace; 1991\textvisiblespace; 1990e ~~\dots\\  
    \midrule
    CASIA & 2 &  \begin{CJK}{UTF8}{min} 
        ”。; 
        ”，; 
        er; 
        ”他; 
        -{}-; 
        ”的; 
        le; 
        明，; 
        li; 
        ut;
    \end{CJK} ~~\dots\\
    \midrule
    ASPEC & 2 &  \begin{CJK}{UTF8}{min} きる; きた; きな; きに; りん; きは; き，; きて; きの; きゅ\end{CJK} ~~\dots\\
    \midrule
    \multirow{5}{*}{WMT} & 2 & e\textvisiblespace; s\textvisiblespace; \textvisiblespace d; t\textvisiblespace; \textvisiblespace l; es; on; \textvisiblespace a; de; en ~~\dots\\
    & 3 & übe; Rüc; rüb; öve; ürs; Köp; üsl \\
    & 4 & ümov; ölln; rüng; Jürg; ülle; Müsl; 
  Müni; üric; üdig; ürri ~~\dots\\
    & 5 & iñera; Mölln; örsdo; hönha\\
    & 6 & ürdo\textvisiblespace d; ñora B \\
\bottomrule
\end{tabular}
}
\caption{Up to 10 most frequent source segments for a given length for the \ACTIS on the dev sets.}
\label{tab:segs}
\end{table*} 

For WMT observe the following patterns (Table~\ref{tab:segs}): identified character 2-gram segments are
all very frequent in the training data (8\=/11M occurrences) while longer
segments are very rare (max.\ 1k occurrences) or completely absent from the
training data; higher order segments include umlauts (ü, ö), which are parts of
the vocabulary, but are atypical for French, except for loan words or proper names
in German, which should be treated as one unit semantically. As for IWSLT, we
observe that both very frequent and very rare patterns constitute segments.

\begin{figure}[!h]
    \centering
  
    \includegraphics[width=1\columnwidth]{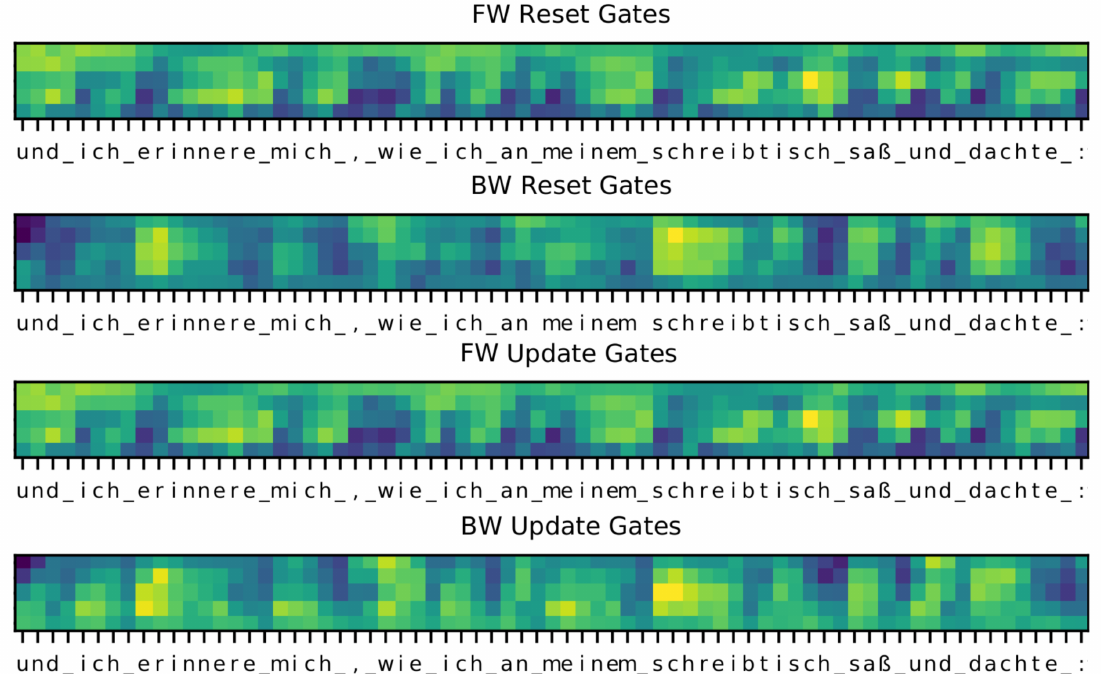} 
    \caption{Mean activations for reset and update GRU gates for an IWSLT sentence and the 5-layer character model. The sentence is cropped to a maximum length of 80. Special characters: $\#$ stands for padding, $|$ for end of sentence.
    Dark blue: low values close to 0, bright yellow: high values close to 1.\label{fig:gru-segment5}}
\end{figure}

\paragraph{Gating Behavior.}
To investigate the reasons for success of the deep character-based
encoders and their better or on-par performance with the segmenting ACT-ENC model, we analyzed average activations of GRU gates. 
A GRU cell computes the next state as:
   $s_t = z \odot\tanh (x_t W_{s} + (s_{t-1} \odot r)W_{g})
       + (1-z) \odot s_{t-1}$, 
where $z$ is the update gate and $r$ the reset gate, both being outputs of
sigmoid layers receiving $x_t$ and $s_{t-1}$~\citep{ChoETAL:14}.  Taking a closer look at the
average values of these gates, we find patterns of segmentation as depicted in
Figure~\ref{fig:gru-segment5} for a 5-layer character model.  Most of the time, a whitespace character
triggers a visible change of gate behavior: Forward reset gates close (reset) one character after a whitespace and
backward reset gates close at whitespaces and then both open at the subsequent
character. The update gates show similar regularities, but here the average gate values
are less extreme.  For longer words all gate activations progressively decay
with the length (as also observed for attention in Figure~\ref{fig:att-char5} in Appendix~C).
In addition, the block-wise processing of the compound
`schreibtisch' (German: `writing table') that was correctly split into
`schreib' and `tisch', points to decompounding abilities that pure character-level models possesses beyond simple
whitespace tokenization.  
The pattern for the 1-layer character model is similar (see Figure~\ref{fig:gru-segment1} in Appendix~B), 
compared to which here the forward update gate gets repurposed, focusing only on the first character, which relates to the
attention behavior we also observe in Figure~\ref{fig:att-char1} in Appendix~C.

Overall, this illustrates that the recurrent gates equip pure character models with the capacity to implicitly model input segmentations, 
which would explain why \ACTIS could not find a radically different or advantageous segmentation.

\section{Summary \& Conclusion} \label{sec:summary}

    We proposed an approach to learning 
    (dynamic and adaptive) input and output segmentations 
    for NMT by extending the Adaptive
    Computation Time paradigm by \citet{Graves:16}. Experiments on four
    translations tasks showed that our model prefers to operate on (almost) 
    character level. This is echoed by the quantitative success of purely
    character-level models 
    and a qualitative analysis of gating
    and attention mechanisms, suggesting that our adaptive model rediscovers
    the segmenting capacity already present in gated recurrent, pure character-based models.
    Given this and the absence of many development hurdles with character-based
    models (pipelines, tokenization, hyperparameters), their lower memory
    consumption and higher robustness, the presented dynamic segmentation
    capacity, being primarily a diagnostic research tool, does not seem to be
    necessary to be modelled explicitly.  We hope these insights
    can serve as justification for intensification of research in pure
    character-level NMT models.

\clearpage
\bibliographystyle{acl_natbib_nourl}
\bibliography{references}

\clearpage
\appendix

\section{Segmenting Decoder}\label{sec:decoder}

After having received the \ACTIS's output sequence
$\oo$, i.e. `embeddings' of character blocks, an NMT encoder-decoder can encode it and decode into a sequence of 
hidden states as usual.  While on the input side the \ACTIS and the upstream
NMT layers are simply stacked onto each other, on the output side the NMT
decoder's layers have to be interleaved because of the auto-regressive
processing and teacher-forcing.  Here we describe our implementation
geared towards a standard RNN-based NMT decoder, but the model can be easily
adapted to other architectures, CNN or Transformer.  

\begin{figure}[ht]
\centering
\includegraphics[width=0.25\textwidth]{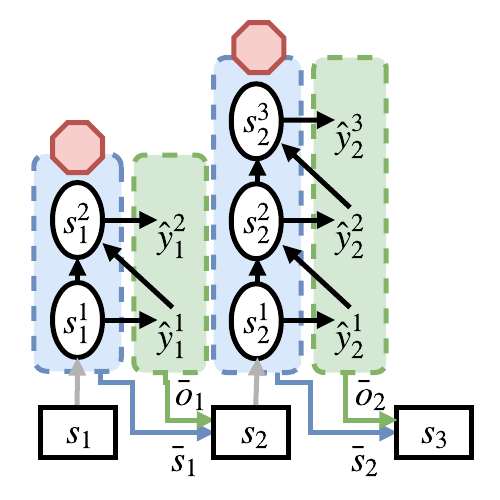}
\caption{Diagram of the \ACTOS decoder.}
\label{fig:act-dec-sketch}
\end{figure}

The \ACTOS predicts one
target character at a time, the halting unit dictates how many of them per segment.
The history input for the RNN state hence consists of a summary of characters
of the previous segment, and the history input for the \ACTOS always consists
of the single previous character.  Figure~\ref{fig:act-dec-sketch} depicts the
\ACTOS decoder.

\begin{algorithm}[!h]
\begin{algorithmic}[1]
\Require{Parameters $W_{o},b_{o},W_{h},b_{h}$, function $\mathcal{S}$, output embeddings $E_{trg}$, decoder RNN $\mathcal{R}$, attention mechanism $\mathcal{A}$, RNN to ACT projection $\mathcal{P}$, output layer $\mathcal{O}$, initial decoder state $s_0$}
\Ensure{Penalty $R$, output sequence $\hat{Y}$}
    \State{$\hat{Y} = [\;] $}\Comment{empty sequence} 
    \State{$R = 0$} \Comment{init remainder}
    \State{$e_1 = E_{trg} \langle\text{bos}\rangle$} \Comment{initial embedded input}
\For{$t = 1 \dots T_{max}$} \Comment{RNN loop}
	\If{$\langle\text{eos}\rangle \in \hat{Y}$} break\label{actos:breakrnn}
	\EndIf
	\State{$c_t = \mathcal{A}(s_{t-1})$} \Comment{attention vector} \label{actos:attend}
	\State{$s_t = \mathcal{R}(s_{t-1},[e_{t}; c_t])$} \Comment{new RNN state}
	\State{$\bar{s} = \bar{o} = \mathbf{0}$} \Comment{init state and output averages}
	\State{$\bar{h} = 0$} \Comment{init halting sum}
	\State{$s_t^0 = \mathcal{P}(s_t)$} \Comment{init ACT state}
	\State{$i_t^0 = e_t$} \Comment{init ACT input}
	\For{$n = 1 \dots T_{max}$} \Comment{ACT loop}
		\State{$s_t^n = \mathcal{S}(s_{t}^{n-1}, i_t^{n-1})$} \label{actos:state} \Comment{new ACT state}
		\State{$o_t^n = W_{o}s_t^n + b_{o}$} 	 \label{actos:out} \Comment{new ACT output}
		\State{$\hat{y}_t^n = \text{arg}\max{\mathcal{O}(o_t^n)}$} \Comment{greedy prediction} \label{actos:greedy}
        \State{$\hat{Y} = \hat{Y}^\frown [\hat{y}_t^n]$} \Comment{append}\label{actos:expand} 
		\State{$h_t^n = \sigma(W_{h} s_t^n + b_{h})$} \label{actos:score} \Comment{halt. score}
        \State{$f = [\![\bar{h}+h_t^n \geq 1-\epsilon ]\!]$} \Comment{halting flag}
        \State{$H = H + h_t^n$} \Comment{increment halting sum}
        \State{$p_t^n = (1-f) \, h_t^n + f \, (1-H)$} \label{actos:prob} \Comment{halt. prob.}
		\State{$\bar{s} = \bar{s} + p_t^n s_t^n$} \Comment{update averages}
        \State{$\bar{o} = \bar{o} + p_t^n E_{trg}\hat{y}_t^n$} \label{actos:meano} \Comment{update output}
		\State{$R = R + (1-f)\,h_t^n$} \label{actos:penalty} \Comment{increment penalty}
		\State{$i_t^{n} = E_{trg}\hat{y}_t^n$} \Comment{next ACT step input} \label{actos:prepinput}
		\If{$f$ \textbf{or} $\langle\text{eos}\rangle \in \hat{Y}$} break \label{actos:halt} 
		\EndIf
   	\EndFor
   \State{$s_t = \bar{s}$} \Comment{next RNN state history}
   \State{$e_t = \bar{o}$} \Comment{next RNN input} \label{actos:meanemb}
   \State{$R = \frac{1-R}{t}$} \Comment{compute total penalty}
\EndFor

\end{algorithmic}
\caption{\ACTOS}
\label{alg:actos}
\end{algorithm}

Algorithm \ref{alg:actos} describes the ACT computations on the decoder side during inference.\footnote{For simplicity we use the same names for ACT parameters as in \ACTIS. In practice they are not the same weights but can be shared if their sizes agree.} While the elementary ACT computations, such as the computation of the ACT state, halting probabilities, remainder and penalty (lines \ref{actos:state}, \ref{actos:score}, \ref{actos:penalty}) are the same as in \ACTIS (cf.\ Algorithm \ref{alg:actis}), additional complexity is introduced by the fact that the history is computed on the fly. Usually the input to each attention RNN step is the embedded previously generated target symbol, but with ACT we generate an arbitrary number of characters in each RNN step, such that the history is instead a weighted mean over embedded generated symbols, i.e.\ a summary of the previously generated segment (line \ref{actos:meanemb}). The attention\footnote{The attention vector can involve a complex computation, as e.g.\ in (4) of \cite{Sockeye:17}.} is only computed on the RNN level (line \ref{actos:attend}) with the rationale that alignments are modelled between segments, such that each element within one segment attends to the same source.\footnote{In the ACT state computation, the attention vector is not fed as input (line \ref{actos:state}) but only to the computation of the initial ACT state. It might improve the model if the attention vector was also fed in every ACT step, since connections to the encoder were shorter.} 

The greedy choice of the generated target symbol (line~\ref{actos:greedy}) can
be replaced by sampling if the training objective requires it (e.g.\ scheduled
sampling \cite{BengioETAL:15} or minimum risk training \cite{ShenETAL:16}).

For standard cross-entropy training the target history in \ACTOS $\hat{y}^n_t$
(lines \ref{actos:meano} and \ref{actos:prepinput}) is replaced by the target
symbols at the corresponding positions $y_t^n$ (teacher forcing).  Similar to
\ACTIS the \ACTOS penalty is added to the training loss with coefficient
$\tau$:\begin{align} L_{\text{\ACTIS-DEC}} = L_{\text{\ACTIS}} + \tau\, R.
\end{align} A larger $\tau$ will again prefer smaller $R$, i.e.\ smaller
remainders with result from shorter segments.

\subsubsection{Limitations.}
Since in every decoder step an arbitrary number of characters can be generated
the comparison of beam search hypotheses becomes hard and its implementation
non-trivial.  Another challenge is efficient minibatching: Due to the flexible
breaking conditions in the nested loops (line~\ref{actos:breakrnn}
and~\ref{actos:halt}) and therefore variable output lengths, there can be a lot of overhead in computation time for each batch. 
When working with minibatches,
the breaks are only executed as soon as every element of the batch fulfills the
condition. During training, when the reference output length is known, there
are two extreme cases that can occur in the same batch: 1) full number of RNN
steps required, i.e.\ \ACTOS halts after every step, 2) full number of \ACTOS
steps required, i.e.\ \ACTOS does not halt. In practice, however, we observe
roughly coordinated halting behaviour for instances in the same batch, which
gives reason to believe that the worst case scenario is rare.

\section{Gating Behaviour for One-Layer Character Models}\label{sec:gating}

\begin{figure}
    \centering 
    \includegraphics[width=1\columnwidth]{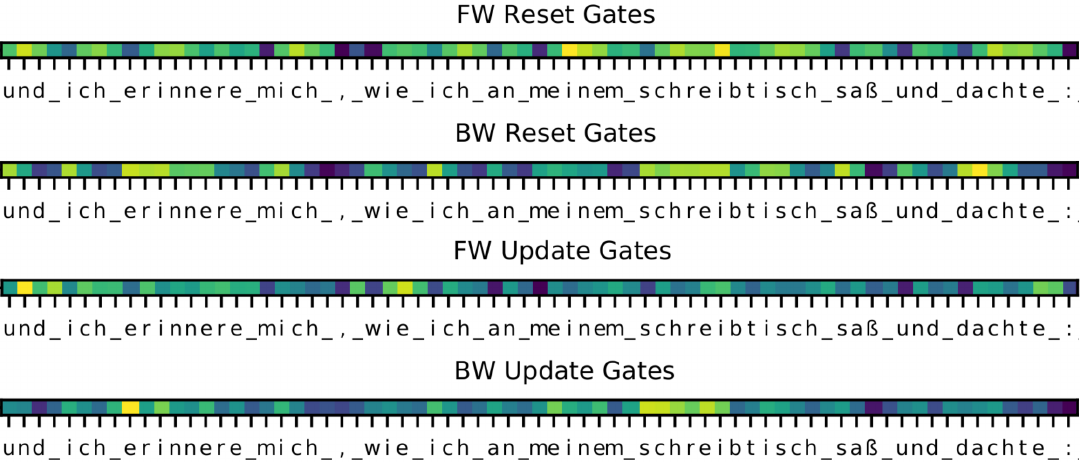}
    \caption{1-layer encoder: Mean activations for reset and update GRU gates for an IWSLT sentence. The sentence is cropped to a maximum length of 80. Special symbols: $\#$ stands for padding, $|$ for end of sentence. Dark blue: low values close to 0, bright yellow: high values close to 1.}
    \label{fig:gru-segment1}
\end{figure}

We plot the average update and reset gates activations for a single layer
character model in Figure~\ref{fig:gru-segment1}.  As for the case of 5-layer model
(Figure~\ref{fig:gru-segment5}) we also observe large changes of their
amplitude on whitespaces and punctuation, and on German compound words. 

\flushcolsend

\clearpage
\onecolumn
\section{Attention Plots}\label{sec:attention}
\vspace{-1.5em}
\begin{figure}[!h]
\centering
    \subfloat[1 layer\label{fig:att-char1}]{\includegraphics[width=0.63\textwidth]{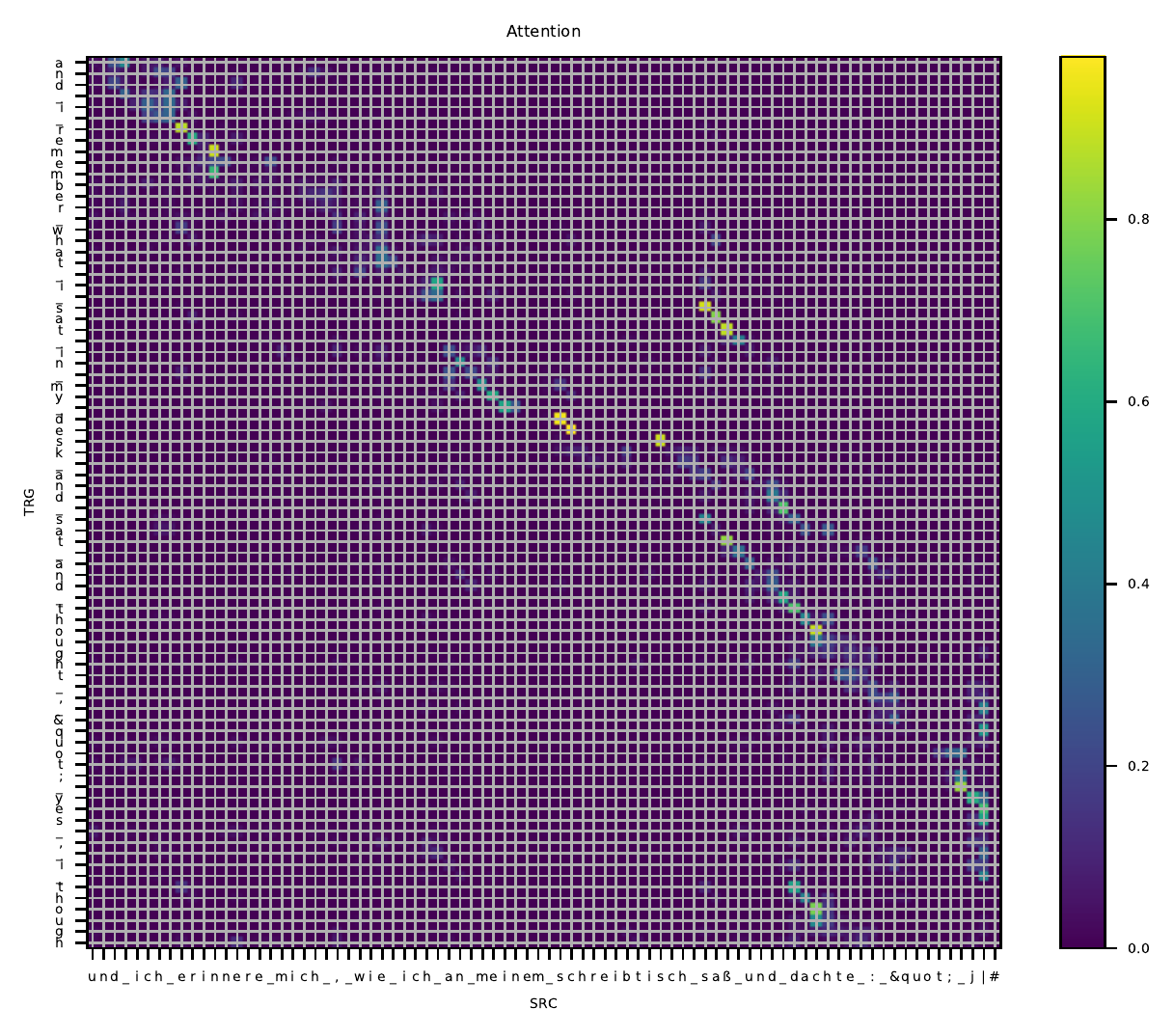}
}\\
    \subfloat[5 layers\label{fig:att-char5}]{\includegraphics[width=0.63\textwidth]{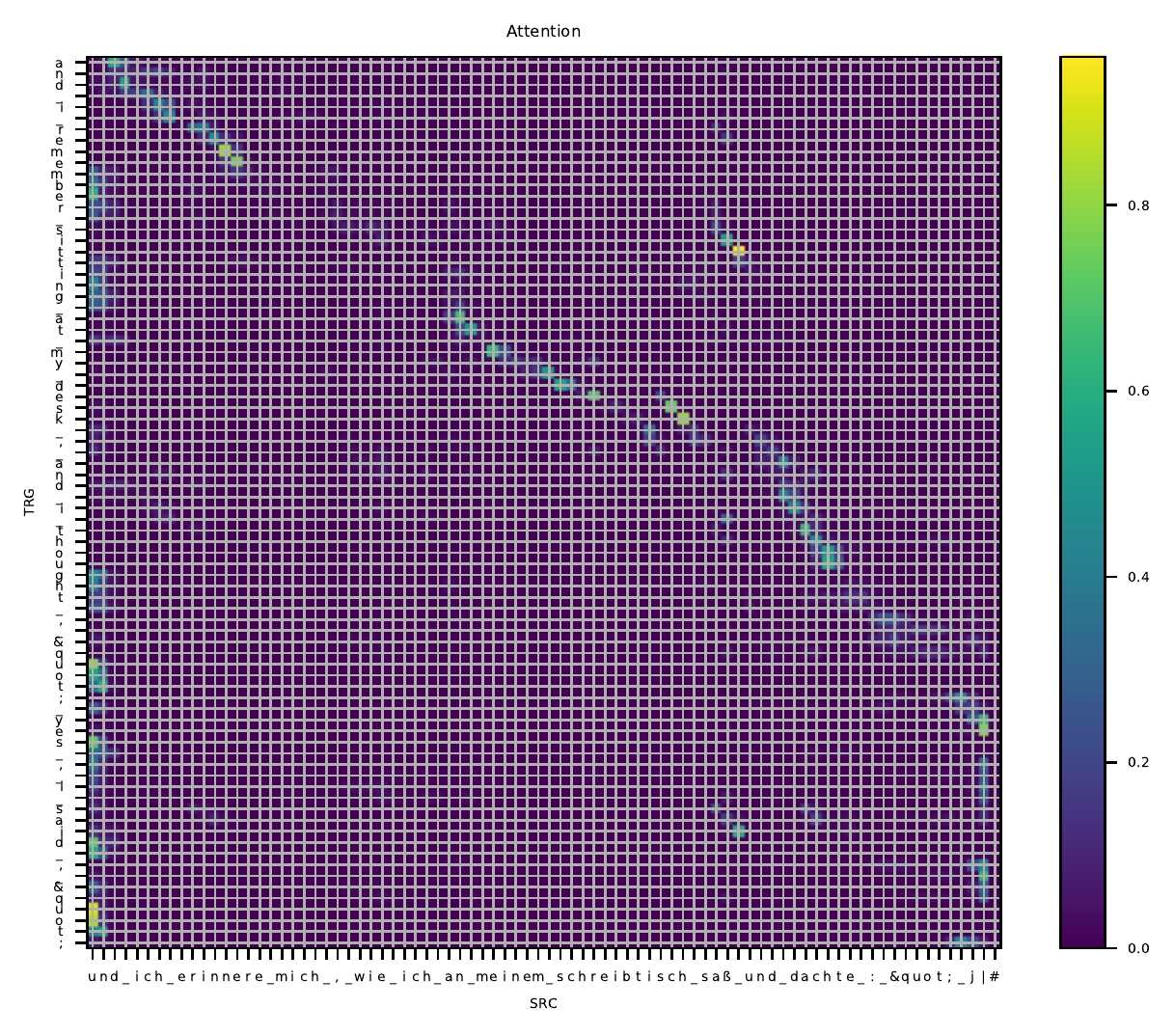}
}
\caption{Attention scores for 1 and 5-layer encoder character-based models. The sentences are cropped to a maximum length of 80. 
    Special symbols: $\#$ stands for padding, $|$ for end of sentence.
    }
\end{figure}

\newpage
\section{Translation Examples on ASPEC and WMT}\label{sec:examples}

Table~\ref{tab:examples-aspec} lists examples for
BPE and character-level models for ASPEC. In the first example, both BPE and
character-based models struggle with the first part of the translation
(`second' vs. `middle ' vs. `medium') and the long noun-phrase at the
end.  
In both examples the \ACTIS system suffers from repetitions of
phrases.

Table~\ref{tab:examples-wmt} presents examples from
the WMT dev set for different models. The first example's source is
incomplete with a missing last word and a period (they appear in the reference). 
The word-based models shows no hallucination behavior, while
BPE- and character-based models make up additional words (`the company\&apos;s
second stage', `people') or mistranslate verbs (`gone', `gained',
`fame').  The missing character in `orgnisés' in the second example is an
interesting showcase that contributes to our advocacy of using character-based models: Both
character-based and our \ACTIS models manage to correct this typo due to their strong
language-modeling abilities (that were somewhat detrimental for IWSLT in Table~\ref{tab:examples-iwslt}). All
models have difficulties with the rare `clou', being translated as $<$unk$>$,
`bell', `club' or `cloud'. While the word-based model can only output
$<$unk$>$, the subword models try to find or translate a word that it similar
to `clou' (French for `bell' is `cloche'). 

\begin{table*}[h]
\resizebox{\textwidth}{!}{
\begin{tabular}{ll}
\toprule
\textbf{Ref} & With the second one, sodium hypochlorite is injected for detection of a compound and measurement of gas produced by reaction with it. \\
\midrule
\multirow{2}{*}{\textbf{WP}} & \begin{CJK}{UTF8}{min} \_ 中 者 では ， 次 亜 塩素 酸 ナトリウム を注入し ， 生成 成分 の検出 とその 反応 に伴う ガス の測定 を行う 。\end{CJK}\\
    & In the middle part, sodium hypochlorite is injected, and the detection of the generated component and the gas with the reaction are carried out.\\
\midrule
\multirow{2}{*}{\textbf{\ACTIS}} & \begin{CJK}{UTF8}{min}中$|$者$|$で$|$は$|$，$|$次$|$亜$|$塩$|$素$|$酸$|$ナ$|$ト$|$リ$|$ウ$|$ム$|$を$|$注$|$入$|$し$|$，$|$生$|$成$|$成$|$分$|$の$|$検$|$出$|$と$|$そ$|$の$|$反$|$応$|$に$|$伴$|$う$|$ガ$|$ス$|$の$|$測$|$定$|$を$|$行$|$う$|$。$|$ \end{CJK}\\ 
& In the medium with the sodium chloride, the detection of generation and the reaction with the reaction was measured with the detection of the formatio\\
\midrule
\multirow{2}{*}{\textbf{Char}} & \begin{CJK}{UTF8}{min}中$|$者$|$で$|$は$|$，$|$次$|$亜$|$塩$|$素$|$酸$|$ナ$|$ト$|$リ$|$ウ$|$ム$|$を$|$注$|$入$|$し$|$，$|$生$|$成$|$成$|$分$|$の$|$検$|$出$|$と$|$そ$|$の$|$反$|$応$|$に$|$伴$|$う$|$ガ$|$ス$|$の$|$測$|$定$|$を$|$行$|$う$|$。$|$ \end{CJK}\\
    & In the middle of the method, sodium hypochlorite is injected and the detection of the production component and the measurement of gas with the\\& reaction are carried out. \\
\midrule
\midrule
    \textbf{Ref} & In case of the beam subjected to axial tensile force, the shear crack position and its angle can be changed by the size of axial force.\\
\midrule
\multirow{2}{*}{\textbf{WP}} & \begin{CJK}{UTF8}{min} \_ 軸方向 引張 力
を受ける 梁 の場合 ， せん断 ひび割れ 位置 及びその 角度 は 軸力 の大きさ
によって 変化する 。 \end{CJK}\\
& In the case of beam subjected to axial tensile force, the crack opening
    position and its angle changes with the axial force.  \\
\midrule
\multirow{2}{*}{\textbf{\ACTIS}} & \begin{CJK}{UTF8}{min}軸$|$方$|$向$|$引$|$張$|$力$|$を$|$受$|$け$|$る$|$梁$|$の$|$場$|$合$|$，$|$せ$|$ん$|$断$|$ひ$|$び$|$割$|$れ$|$位$|$置$|$及$|$び$|$そ$|$の$|$角$|$度$|$は$|$軸$|$力$|$の$|$大$|$き$|$さ$|$に$|$よ$|$っ$|$て$|$変$|$化$|$す$|$る$|$。$|$ \end{CJK}\\ 
& In the case of beam case of axial tension, the shear strain convection position and the angle of the axial force changes with the size of the axial forc\\ 
\midrule
\multirow{2}{*}{\textbf{Char}} & \begin{CJK}{UTF8}{min}軸$|$方$|$向$|$引$|$張$|$力$|$を$|$受$|$け$|$る$|$梁$|$の$|$場$|$合$|$，$|$せ$|$ん$|$断$|$ひ$|$び$|$割$|$れ$|$位$|$置$|$及$|$び$|$そ$|$の$|$角$|$度$|$は$|$軸$|$力$|$の$|$大$|$き$|$さ$|$に$|$よ$|$っ$|$て$|$変$|$化$|$す$|$る$|$。$|$ \end{CJK}\\
& In the case of a beam which received axial tension, the shear crack position and their angle varies with the size of the axial force.\\
\bottomrule
\end{tabular}
}
\caption{Examples from the ASPEC dev set: segmented sources and greedy translations. Word, BPE and \ACTIS models have 1 encoder layer, and the character model has 2 layers.}
\label{tab:examples-aspec}
\end{table*}

\begin{table*}[h]
\resizebox{\textwidth}{!}{
\begin{tabular}{ll}
\toprule
    \textbf{Ref} & In Montenegro they won 1:0 and celebrate a 200 million windfall. \\
\midrule
    \multirow{2}{*}{\textbf{Word}} & Ils ont gagné au Monténégro 1 : 0 et fêtent la qualification pour 200 millions\\
& They won in Montenegro 1 : 0 and $<$unk$>$ the qualification for 200 million . \\
\midrule
    \multirow{2}{*}{\textbf{BPE}} & Ils ont gagné au Monténégro 1 : 0 et f@@ ê@@ tent la qualification pour 200 millions\\
& They have gained in Montenegro 1 : 0 and fame the qualification for the 200 million people . \\
\midrule
    \multirow{2}{*}{\textbf{\ACTIS}} & Il$|$s $|$on$|$t $|$ga$|$gn$|$é $|$au$|$ M$|$on$|$té$|$né$|$gr$|$o $|$1 $|$:$|$ 0$|$ e$|$t $|$f$|$ê$|$t$|$en$|$t $|$la$|$ q$|$ua$|$li$|$f$|$ic$|$at$|$
                         io$|$n $|$p$|$ou$|$r $|$20$|$0 $|$mi$|$ll$|$io$|$ns$|$\\ 
& They have gone to Montenegro 1 : 0 and celebrating the qualification for 200 million \\
\midrule
    \multirow{2}{*}{\textbf{Char}} & I$|$l$|$s$|$ $|$o$|$n$|$t$|$ $|$g$|$a$|$g$|$n$|$é$|$ $|$a$|$u$|$ $|$M$|$o$|$n$|$t$|$é$|$n$|$é$|$g$|$r$|$o$|$ $|$1$|$ $|$:$|$ $|$0$|$ $|$e$|$t$|$ $|$f$|$ê$|$t$|$e$|$n$|$t $|$l$|$a$|$ $|$q$|$u$|$a$|$l$|$i$|$f$|$i$|$c$|$a$|$t$|$i$|$o$|$n$|$ $|$p$|$o$|$u$|$r$|$ $|$2$|$0$|$0$|$ $|$m$|$i$|$l$|$l$|$i$|$o$|$n$|$s$|$\\
& They won the company \&apos;s second stage in Montenegro 1 : 0 and celebrate the qualification for 200 million \\
\midrule
\midrule
    \textbf{Ref} & The main focus of the festival is on two concerts taking place on November 17.\\
\midrule
    \multirow{2}{*}{\textbf{Word}} & Le clou du festival est formé de deux concerts orgnisés le 17 novembre .\\
& The $<$unk$>$ of the festival is composed of two concerts on 17 November . \\
 \midrule
    \multirow{2}{*}{\textbf{BPE}} & Le clo@@ u du festival est formé de deux concerts or@@ gn@@ isés le 17 novembre .\\
& The festival \&apos;s bell is composed of two concerts , on 17 November .\\
             \midrule
    \multirow{2}{*}{\textbf{\ACTIS}} & L$|$e c$|$lo$|$u $|$du$|$ f$|$es$|$ti$|$v$|$al$|$ e$|$st$|$ f$|$or$|$mé$|$ d$|$e $|$de$|$ux$|$ c$|$on$|$c$|$er$|$ts$|$ o$|$rg$|$ni$|$sé$|$s $|$le$|$ 1$|$7 $|$no$|$v$|$em$|$b$|$re$|$ .$|$\\ 
& The festival club is the form of two concerts organized on 17 November .\\
               \bottomrule
\end{tabular}
}
\caption{Examples from the WMT dev set: segmented sources and greedy translations. Word, BPE and \ACTIS models have 1 encoder layer, and the character model has 6 layers.}
\label{tab:examples-wmt}
\end{table*}

 \end{document}